\documentclass{article}

\usepackage[accepted]{icml2025} 
\usepackage[utf8]{inputenc}
\usepackage{enumitem}
\usepackage[T1]{fontenc}
\usepackage{hyperref}
\usepackage{url}
\usepackage{booktabs}
\usepackage{amsfonts}
\usepackage{nicefrac}
\usepackage{microtype}
\usepackage{amsmath, amssymb, amsthm, mathtools, bm}
\usepackage{graphicx}
\usepackage{subfigure} 
\usepackage{natbib}
\usepackage{tcolorbox}
\usepackage{xcolor}

\definecolor{princetonorange}{rgb}{1.0, 0.56, 0.0}
\definecolor{deepskyblue}{rgb}{0.0, 0.75, 1.0}
\definecolor{limegreen}{rgb}{0.2, 0.8, 0.2}
\definecolor{mediumpurple}{rgb}{0.58, 0.44, 0.86}
\definecolor{crimsonred}{rgb}{0.86, 0.08, 0.24}

\theoremstyle{plain}
\newtheorem{theorem}{Theorem}[section]

\theoremstyle{definition}
\newtheorem{definition}{Definition}[section]
\newtheorem{assumption}{Assumption}[section]

\theoremstyle{remark}

\newcommand{\R}{\mathbb{R}}
\newcommand{\N}{\mathbb{N}}
\newcommand{\norm}[1]{\left\lVert#1\right\rVert}

\newcommand{\diff}{\,\mathrm{d}}

\icmltitlerunning{A Statistical Physics of Language Model Reasoning}

\title{A Statistical Physics of Language Model Reasoning}

\author{%
  \textbf{Jack David Carson} \\
  Massachusetts Institute of Technology\\
  \texttt{jdcarson@mit.edu} \\
\and
  Amir Reisizadeh \\
  Massachusetts Institute of Technology\\
  \texttt{amirr@mit.edu} \\
}
\begin{document}

\maketitle


\icmlcorrespondingauthor{Jack David Carson}{jdcarson@mit.edu}

\icmlkeywords{Stochastic Processes, Transformer Interpretability, Chain-of-Thought Reasoning, Dynamical Systems, Large Language Models}

\begin{abstract}
Transformer LMs show emergent reasoning that resists mechanistic understanding. We offer a statistical physics framework for continuous-time chain-of-thought reasoning dynamics. We model sentence-level hidden state trajectories as a stochastic dynamical system on a lower-dimensional manifold. This drift-diffusion system uses latent regime switching to capture diverse reasoning phases, including misaligned states or failures. Empirical trajectories (8 models, 7 benchmarks) show a rank-40 projection (balancing variance capture and feasibility) explains ~50\% variance. We find four latent reasoning regimes. An SLDS model is formulated and validated to capture these features. The framework enables low-cost reasoning simulation, offering tools to study and predict critical transitions like misaligned states or other LM failures.
\end{abstract}

\section{Introduction}

Transformer LMs \citep{vaswani2017attention}, trained for next-token prediction \citep{radford2019language, brown2020language}, show emergent reasoning like complex cognition \citep{wei2022chain}. Standard analyses of discrete components (e.g., attention heads \citep{elhage2021mathematical, olsson2022incontext}) provide limited insight into longer-scale semantic transitions in multi-step reasoning \citep{allenzhu2023hierarchical, lopezotal2025linguistic}. Understanding these high-dimensional, prediction-shaped semantic trajectories, particularly how they might cause misaligned states, is a key challenge \citep{li2023emergent, nanda2023emergent}.

We model reasoning as a continuous-time dynamical system, drawing from statistical physics \citep{chaudhuri2016computational, schuecker2018optimal}. Sentence-level hidden states $h(t) \in \R^D$ evolve via a stochastic differential equation (SDE):
\begin{equation}\label{eq:intro_sde}
\diff h(t) = \mu(h(t), Z(t)) \diff t + B(h(t), Z(t)) \diff W(t),
\end{equation}
with drift $\mu$, diffusion $B$, Wiener process $W(t)$, and latent regimes $Z(t)$. This decomposes trajectories into trends and variations, helping identify deviations. As full high-dimensional SDE analysis (e.g., $D>2048$ for most LMs) is impractical, we use a lower-dimensional manifold capturing significant variance for modeling.

This continuous-time dynamical systems perspective offers several benefits: 

\begin{tcolorbox}[colback=gray!5!white, colframe=black!60!white, arc=1.5mm, boxrule=0.8pt,
title=\textcolor{white!75}{\textbf{Core Advantages}}, 
 fonttitle=\bfseries\normalsize, 
left=3mm, right=3mm, top=3mm, bottom=3mm, 
                  nobeforeafter, 
                  boxsep=1mm 
                  ]
\begin{itemize}[itemsep=2pt, topsep=2pt, leftmargin=*, labelsep=3mm,
                  parsep=1pt, 
                  ]
    \item[\textcolor{princetonorange}{$\bullet$}] \textbf{Principled Abstraction}: Enables a mathematically grounded, semantic-level view of reasoning, akin to statistical physics approximations, moving beyond token mechanics for robust interpretation of reasoning pathways and potential misalignments.
    
    \item[\textcolor{deepskyblue}{$\bullet$}] \textbf{Tractable Latent Structure ID}: Makes analysis of reasoning trajectories feasible by focusing on a low-dimensional manifold (e.g., rank-40 PCA capturing ~50\% variance) that describes significant structured evolution.
    
    \item[\textcolor{limegreen}{$\bullet$}] \textbf{Reasoning Regime Discovery}: Uncovers distinct latent semantic regimes with unique drift/variance profiles, suggesting context-driven switching and offering insight into how models might slip into different reasoning states (Appx. E).
    
    \item[\textcolor{mediumpurple}{$\bullet$}] \textbf{Efficient Surrogate Model}: Our SLDS accurately models and reconstructs reasoning trajectories with significant computational savings, facilitating the study of how reasoning processes unfold.
    
    \item[\textcolor{crimsonred}{$\bullet$}] \textbf{Failure Mode Analysis}: Provides tools to study critical transitions, robustness, and predict inference-time failure modes or misaligned states in LLM reasoning.
\end{itemize}
\end{tcolorbox}

Chain-of-thought (CoT) prompting \citep{wei2022chain, wang2023understanding} has demonstrated that LMs can follow structured reasoning pathways, hinting at underlying processes amenable to a dynamical systems description. While prior work has applied continuous-time models to neural dynamics generally, the explicit modeling of transformer reasoning at these semantic timescales, particularly as an approximation for impractical full-dimensional analysis, has been largely unexplored. Our work bridges this gap by pursuing an SDE-based perspective informed by empirical analysis of transformer hidden-state trajectories.

This paper is structured as follows: Section~\ref{sec:preliminaries} introduces the mathematical formalism of SDEs and regime switching. Section~\ref{sec:data_empirical_decomp} details our data collection and initial empirical findings that motivate the model, including the practical need for dimensionality reduction. Section~\ref{sec:slds_definition} formally defines the SLDS model. Section~\ref{sec:experiments_validation} presents experimental validation, including model fitting, generalization, ablation studies, and a case study on modeling adversarial belief shifts as an example of predicting misaligned states.

\section{Mathematical Preliminaries}
\label{sec:preliminaries}

We conceptualize the internal reasoning process of a transformer LM as a continuous-time stochastic trajectory evolving within its hidden-state space. Let $h_t \in \R^D$ be the final-layer residual embedding extracted at discrete sentence boundaries $t=0,1,2,\dots$. To capture the rich semantic evolution across reasoning steps, we treat these discrete embeddings as observations of an underlying continuous-time process $h(t): \R_{\ge 0} \to \R^D$. The direct analysis of such a process in its full dimensionality (e.g., $D \ge 2048$) is often computationally prohibitive. We therefore aim to approximate its dynamics using SDEs, potentially in a reduced-dimensional space.

\begin{definition}[Itô SDE]\label{def:ito_sde}
An Itô stochastic differential equation on the state space $\R^D$ is given by:
\begin{equation}\label{eq:ito_sde_main}
    \diff h(t) = \mu(h(t))\diff t + B(h(t))\diff W(t),\quad h(0) \sim p_0,
\end{equation}
where $\mu:\R^D\to\R^D$ is the deterministic \emph{drift} term, encoding persistent directional dynamics. The matrix $B:\R^D\to\R^{D\times D'}$ is the \emph{diffusion} term, modulating instantaneous stochastic fluctuations. $W(t)$ is a $D'$-dimensional Wiener process (standard Brownian motion), and $p_0$ is the initial distribution. The noise dimension $D'$ can be less than or equal to the state dimension $D$.
\end{definition}

The drift $\mu(h(t))$ represents systematic semantic or cognitive tendencies, while the diffusion $B(h(t))$ accounts for fluctuations due to local uncertainties, token-level variations, or inherent model stochasticity. Standard conditions ensure the well-posedness of such SDEs:

\begin{theorem}[Well-Posedness {\citep{oksendal2003stochastic}}]
\label{thm:well_posedness}
If $\mu$ and $B$ satisfy standard Lipschitz continuity and linear growth conditions (see Appendix A), the SDE
\begin{equation}\label{eq:basic_sde_main}
\diff h(t) = \mu(h(t))\diff t + B(h(t))\diff W(t)
\end{equation}
has a unique strong solution for a given $D'$-dimensional Wiener process $W(t)$.
\end{theorem}

We focus on dynamics at the sentence level:
\begin{definition}[Sentence-Stride Process]
\label{def:sentence_stride}
The \emph{sentence-stride} hidden-state process is the discrete sequence $\{h_t\}_{t\in\N}$ obtained by extracting the final-layer transformer state immediately following each detected sentence boundary. This emphasizes mesoscopic, semantic-level changes over finer-grained token-level variations.
\end{definition}

To analyze these dynamics in a computationally manageable way, particularly given the high dimensionality $D$ of $h(t)$, we utilize projection-based dimensionality reduction. The goal is to find a lower-dimensional subspace where the most significant dynamics, for the purpose of modeling the SDE, unfold.
\begin{definition}[Projection Leakage]
\label{def:projection_leakage}
Given an orthonormal matrix $V_k\in\R^{D\times k}$ (where $V_k^\top V_k=I_k$), the \emph{leakage} of the drift $\mu$ under perturbations $v$ orthogonal to the image of $V_k$ (i.e., $v \perp \mathrm{Im}(V_k)$) is
\[
L_k = \sup_{\substack{x\in\R^D,\, \norm{v}\le\epsilon\\v^\top V_k=0}}
\frac{\norm{\mu(x+v)-\mu(x)}}{\norm{\mu(x)}}.
\]
A small leakage $L_k$ implies that the drift's behavior relative to its current direction is not excessively altered by components outside the subspace spanned by $V_k$, making the subspace a reasonable domain for approximation.
\end{definition}

\begin{assumption}[Approximate Projection Closure for Modeling]
\label{assump:projection_closure}
For practical modeling of the SDE (Eq.~\ref{eq:ito_sde_main}), we assume there exists a rank $k$ (e.g., $k=40$ in our work, chosen based on empirical variance and computational trade-offs) and a perturbation scale $\epsilon>0$ such that $L_k \ll 1$. This allows the approximation of the drift within this $k$-dimensional subspace:
\[
\mu(h(t))\approx V_kV_k^\top\mu(h(t))
\]
holds up to an error of order $O(L_k)$. This assumption underpins the feasibility of our low-dimensional modeling approach, enabling the analytical treatment inspired by statistical physics.
\end{assumption}

Empirical observations of reasoning trajectories suggest abrupt shifts, potentially indicating transitions between different phases of reasoning or slips into misaligned states. This motivates a regime-switching framework:
\begin{definition}[Regime-Switching SDE]\label{def:rs_sde_main}
Let $Z(t)\in\{1,\dots,K\}$ be a latent continuous-time Markov chain with a transition rate matrix $T\in\R^{K\times K}$. The corresponding regime-switching Itô SDE is:
\begin{equation}\label{eq:switching_sde_main}
    \diff h(t) = \mu_{Z(t)}(h(t))\diff t + B_{Z(t)}(h(t))\diff W(t),
\end{equation}
where each latent regime $i \in \{1, \dots, K\}$ has distinct drift $\mu_i$ and diffusion $B_i$ functions. This allows for context-dependent dynamic structures \citep{ghahramani2000variational}, crucial for capturing diverse reasoning pathways.
\end{definition}
These definitions establish the mathematical foundation for our analysis of transformer reasoning dynamics as a tractable approximation of a more complex high-dimensional process.

\section{Data and Empirical Motivation}
\label{sec:data_empirical_decomp}

We build a corpus of sentence-aligned hidden-state trajectories from transformer-generated reasoning chains across a suite of models (Mistral-7B-Instruct \citep{Jiang2023Mistral7B}, Phi-3-Medium \citep{Abdin2024Phi3}, DeepSeek-67B \citep{DeepSeekAI2024DeepSeekLLM}, Llama-2-70B \citep{Touvron2023Llama2}, Gemma-2B-IT \citep{GemmaTeam2024Gemma}, Qwen1.5-7B-Chat \citep{Bai2023Qwen}, Gemma-7B-IT (also \citep{GemmaTeam2024Gemma}), Llama-2-13B-Chat-HF (also \citep{Touvron2023Llama2})) and datasets (StrategyQA \citep{Geva2021StrategyQA}, GSM-8K \citep{cobbe2021gsm8k}, TruthfulQA \citep{Lin2022TruthfulQA}, BoolQ \citep{Clark2019BoolQ}, OpenBookQA \citep{Mihaylov2018OpenBookQA}, HellaSwag \citep{Zellers2019HellaSwag}, PiQA \citep{Bisk2020PIQA}, CommonsenseQA \citep{Talmor2021CommonsenseQA2, Talmor2019CommonsenseQA}), yielding roughly 9,800 distinct trajectories spanning $\sim$40,000 sentence-to-sentence transitions.

\subsection{Sentence-Level Dynamics and Manifold Structure for Tractable Modeling}

First, we confirmed that sentence-level increments effectively capture semantic evolution. Figure~\ref{fig:cdf_and_residuals}(a) compares the cumulative distribution functions (CDFs) of jump norms ($\norm{\Delta h_t}$) at both token and sentence strides. Token-level increments show a noisy distribution skewed towards small values, primarily reflecting syntactic variations. In contrast, sentence-level increments are orders of magnitude larger, clearly indicating significant semantic shifts and validating our choice of sentence-stride analysis. To reduce "jitter" from minor variations, we filtered out transitions below a minimum threshold ($\norm{\Delta h_t}\leq 10$ in normalized units), yielding cleaner semantic trajectories.

To uncover underlying geometric structures that could make modeling tractable, we applied Principal Component Analysis (PCA) \citep{jolliffe2002principal} to the sentence-stride embeddings. We found that a relatively low-dimensional projection (rank $k=40$) captures approximately 50\% of the total variance in these reasoning trajectories (details in Appendix A). While reasoning dynamics occur in a high-dimensional embedding space, this finding suggests that a significant portion of their variance is concentrated in a lower-dimensional subspace. This is crucial because constructing and analyzing a stochastic process (like a random walk or SDE) in the full embedding dimension (e.g., 2048) is often impractical. The rank-40 manifold thus provides a computationally feasible domain for our dynamical systems modeling, not necessarily because the process is strictly confined to it, but because it offers a practical and informative approximation.

\subsection{Linear Predictability and Multimodal Residuals}

To assess the predictive structure of the semantic drift within this tractable manifold, we performed a global ridge regression \citep{hoerl1970ridge}, fitting a linear model to predict subsequent sentence embeddings from previous ones:
\begin{align}
h_{t+1} &\approx A h_t + c, \label{eq:global_linear_approx} \\
(A, c) &= \arg\min_{A,c} \sum_t \|\Delta h_t - (A - I)h_t - c\|^2 + \lambda \|A\|_F^2. \label{eq:global_linear_fit_orig}
\end{align}
Using a modest regularization ($\lambda=1.0$), this global linear model achieved an $R^2 \approx 0.51$, indicating substantial linear predictability in sentence-to-sentence transitions.

\begin{figure}[t]
  \centering
  \subfigure[]{%
    \includegraphics[width=0.85\columnwidth]{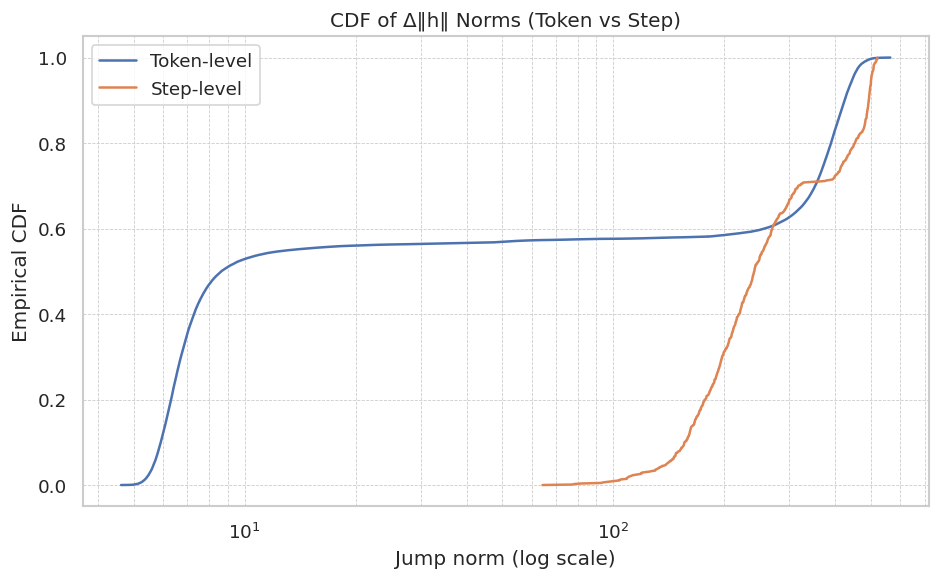} \label{fig:cdf_jump_norms}}
  \subfigure[]{%
    \includegraphics[width=0.85\columnwidth]{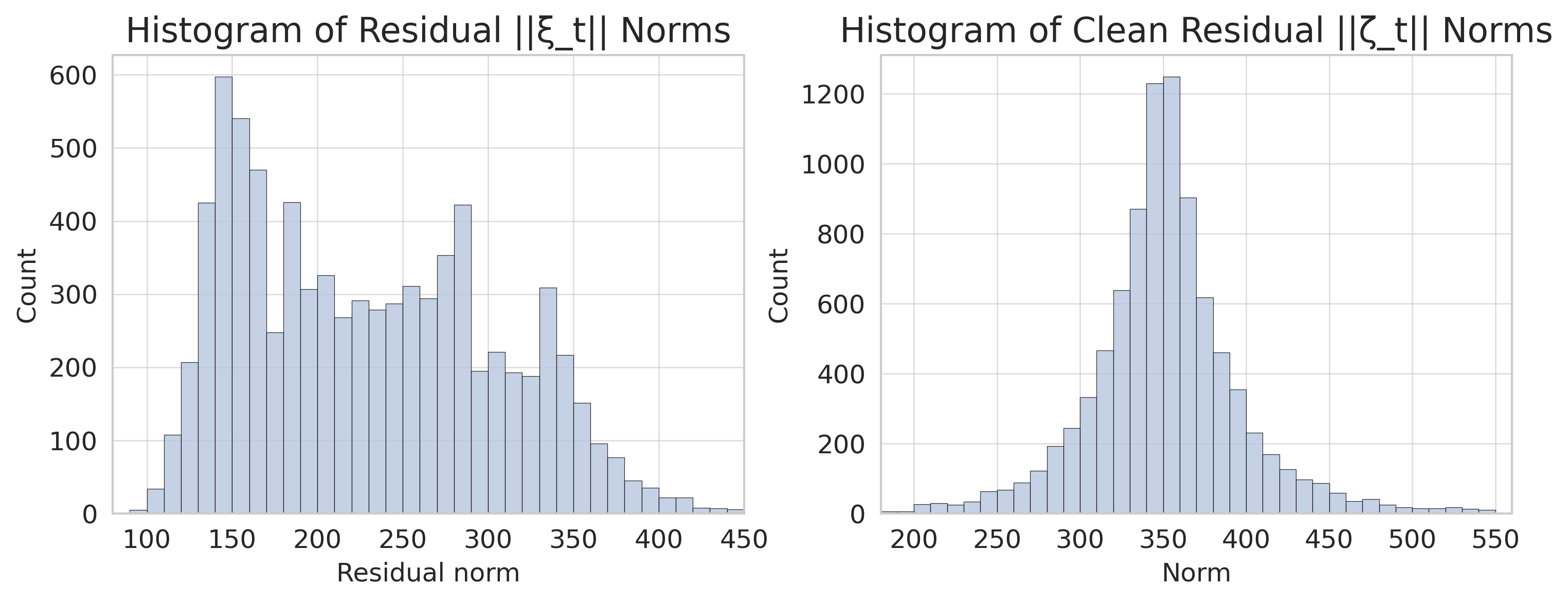} \label{fig:residual_histograms}}
  \caption{(a) CDF comparison of token and sentence jump norms, illustrating that sentence-level increments capture more substantial semantic shifts. (b) Histograms of residual norms from a global linear fit, showing raw residuals $\lVert\xi_t\rVert$ (left) and residuals projected onto a low-rank PCA space $\lVert\zeta_t\rVert$ (right). Both reveal significant multimodality, motivating regime switching to capture distinct reasoning phases or potential misalignments.}
  \label{fig:cdf_and_residuals}
\end{figure}

However, an examination of the residuals from this linear fit, $\xi_t = \Delta h_t - [(A - I)h_t + c]$, revealed persistent multimodal structure, even after the linear drift component was removed (Figure~\ref{fig:cdf_and_residuals}(b)). This multimodality suggests the presence of distinct underlying dynamic states or phases—some potentially representing "misaligned states" or divergent reasoning paths—that are not captured by a single linear model.

Inspired by Langevin dynamics, where a particle in a multi-well potential $U(x)$ can exhibit metastable states (Appendix~E), we interpret these multimodal residual clusters as evidence of distinct latent reasoning regimes. The stationary probability distribution $p_{\mathrm{st}}(x)\propto e^{-U(x)/D}$ for an SDE $\diff x = -U'(x)\diff t + \sqrt{2D}\diff W_t$ becomes multimodal if $U(x)$ has multiple minima and noise $D$ is sufficiently low. Analogously, the observed clusters in our residual analysis point towards the existence of multiple metastable semantic basins in the reasoning process. This strongly motivates the introduction of a latent regime structure to adequately model these richer, nonlinear dynamics and to understand how an LLM might transition between effective reasoning and potential failure modes.

\section{A Switching Linear Dynamical System for Reasoning}
\label{sec:slds_definition}

The empirical evidence that a significant portion of variance is captured by a low-dimensional manifold (making it a practical subspace for analysis, as directly modeling a 2048-dim random walk is often infeasible) and the observation of multimodal residuals motivate a model that combines linear dynamics within distinct regimes with switches between these regimes. Such switches may represent transitions between different cognitive states, some of which could be misaligned or lead to errors.

\subsection{Linear Drift within Regimes}
While a single global linear model (Eq.~\ref{eq:global_linear_approx}) captures about half the variance, the residual analysis (Figure~\ref{fig:cdf_and_residuals}(b)) indicates that a more nuanced approach is needed. We project the residuals $\xi_t$ onto the principal subspace $V_k$ (from Assumption~\ref{assump:projection_closure}, where $k=40$ offers a balance between explained variance and computational cost) to get $\zeta_t = V_k^\top \xi_t$. The clustered nature of these projected residuals $\zeta_t$ suggests that the reasoning process transitions between several distinct dynamical modes or `regimes'.

\subsection{Identifying Latent Reasoning Regimes}
\begin{figure}[t!]
  \centering
  \subfigure[Regime-colored PCA of residuals]{%
    \includegraphics[width=0.43\columnwidth]{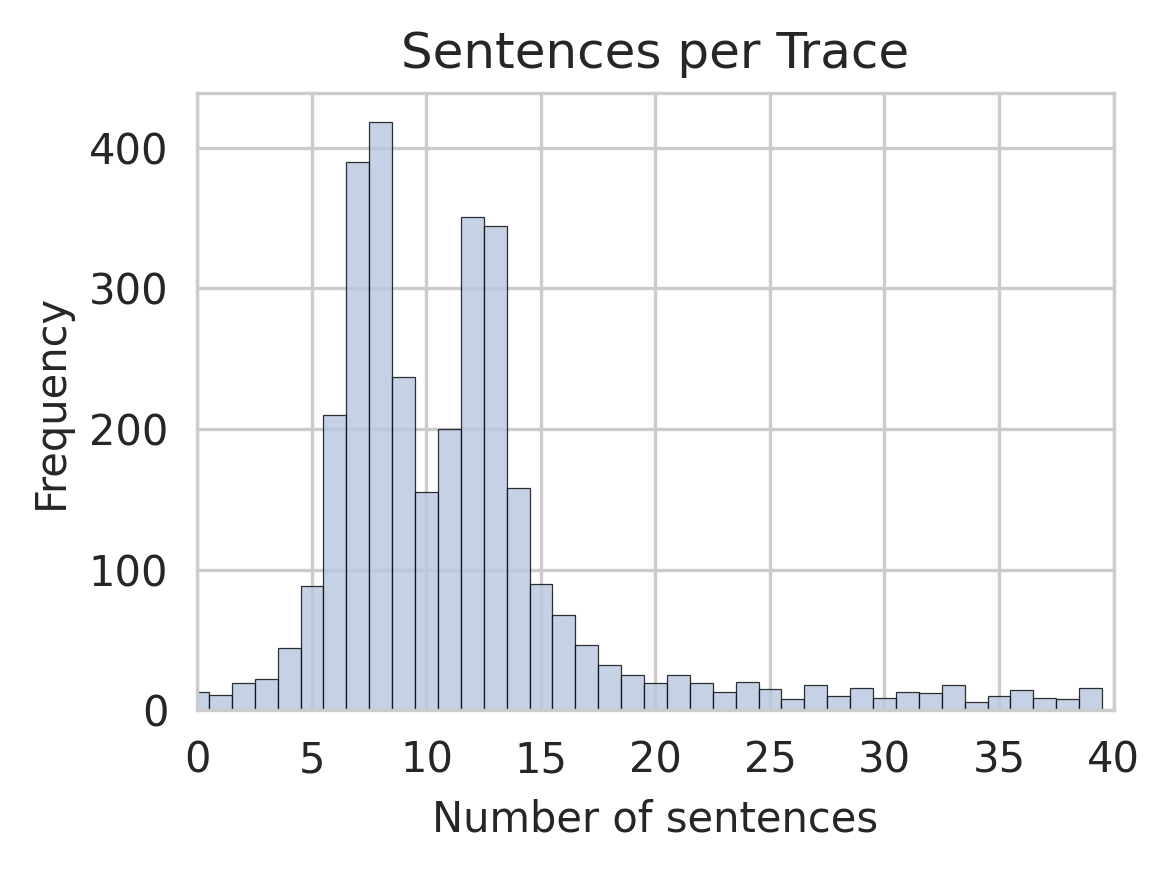} \label{fig:pca_residuals_gmm}}
  \hfill
  \subfigure[Regime-colored histogram of $\norm{\zeta_t}$]{%
    \includegraphics[width=0.43\columnwidth]{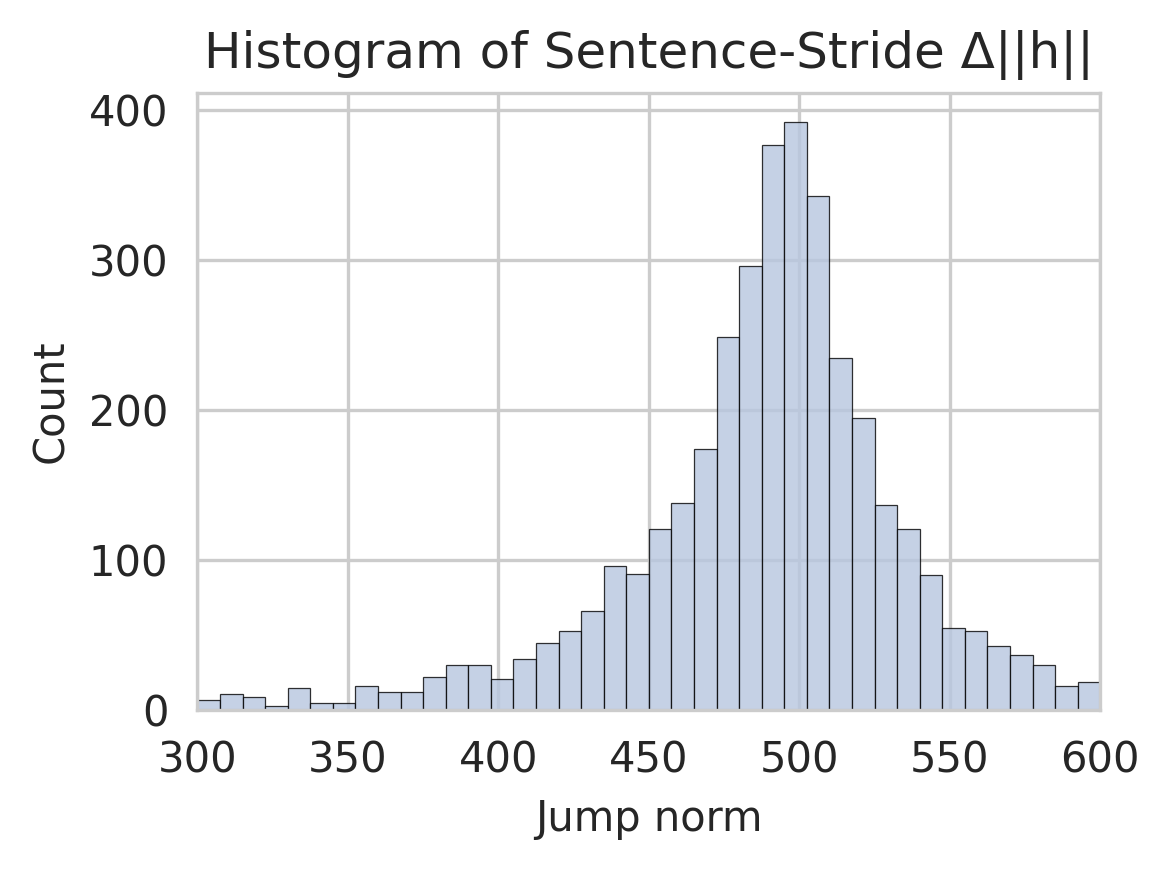} \label{fig:hist_zeta_gmm}}
  \caption{Latent regimes ($K=4$) uncovered by GMM fitting on low-rank residuals $\zeta_t$. (a) Residuals projected onto their first two principal components, colored by GMM assignment, showing distinct clusters. (b) Histogram of residual norms $\norm{\zeta_t}$, colored by GMM regime assignment, further illustrating regime separation. These regimes may capture different reasoning qualities, including potential misalignments.}
  \label{fig:jumps_regimes}
\end{figure}

\begin{figure}[t!]
  \centering
  \includegraphics[width=0.65\columnwidth]{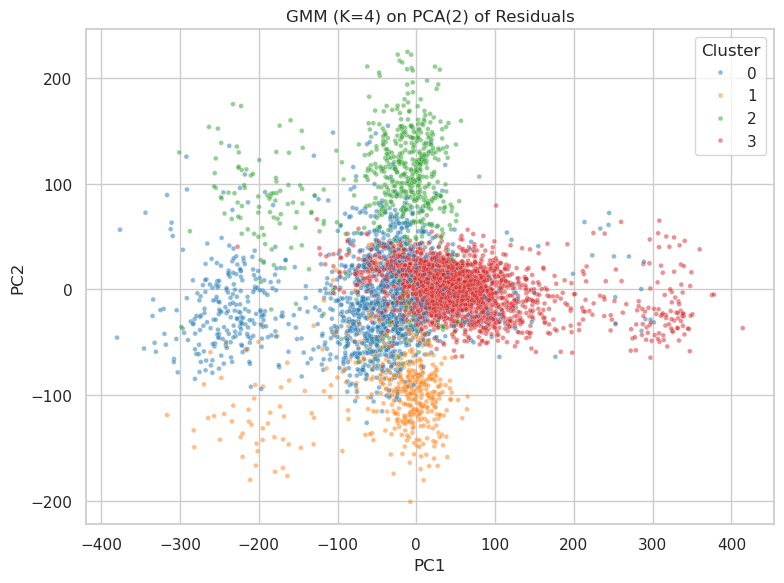}
  \caption{GMM clustering ($K=4$) of low-rank residuals $\zeta_t$, visualized in the space of the first two principal components of $\zeta_t$. The distinct cluster centers provide justification for the regime decomposition, potentially corresponding to different reasoning states or failure modes.}
  \label{fig:regime-scatter_main}
\end{figure}

To formalize these distinct modes, we fit a $K$-component Gaussian Mixture Model (GMM) to the projected residuals $\zeta_t$, following classical regime-switching frameworks \citep{hamilton1989}:
\begin{equation}\label{eq:gmm_residuals}
p(\zeta_t) = \sum_{i=1}^K \pi_i\,\mathcal{N}(\zeta_t\mid\mu_i,\Sigma_i).
\end{equation}
Information criteria (BIC/AIC) suggest $K=4$ as an appropriate number of regimes for our data. While the true underlying multimodality is complex across many dimensions (see Figure~\ref{fig:violin_residuals}, Appendix A), a four-regime model provides a parsimonious yet effective way to capture key dynamic behaviors, including those that might represent misalignments or slips into undesired reasoning patterns, while maintaining computational tractability. We interpret these $K=4$ modes as distinct reasoning phases, such as systematic decomposition, answer synthesis, exploratory variance, or even failure loops, each characterized by specific drift perturbations and noise profiles. Figure~\ref{fig:jumps_regimes} and Figure~\ref{fig:regime-scatter_main} visualize these uncovered regimes in the low-rank residual space.

\subsection{The Switching Linear Dynamical System (SLDS) Model}
We integrate these observations into a discrete-time Switching Linear Dynamical System (SLDS). Let $Z_t \in \{1, \dots, K\}$ be the latent regime at step $t$. The state $h_t$ evolves according to:
\begin{align}
  Z_t &\sim \mathrm{Categorical}(\pi), \quad P(Z_{t+1}=j \mid Z_t=i)=T_{ij},\nonumber\\
  h_{t+1} &= h_t + V_k\bigl(M_{Z_t}(V_k^\top h_t)+b_{Z_t}\bigr) + \varepsilon_t,\label{eq:slds_disc_main}\\
  \varepsilon_t &\sim\mathcal{N}(0,\Sigma_{Z_t}).\nonumber
\end{align}
Here, $M_i\in\R^{k\times k}$ and $b_i\in\R^k$ are the regime-specific linear transformation matrix and offset vector for the drift within the $k$-dimensional semantic subspace defined by $V_k$. $\Sigma_i$ is the regime-dependent covariance for the noise $\varepsilon_t$. The initial regime probabilities are $\pi$, and $T$ is the transition matrix encoding regime persistence and switching probabilities. This SLDS framework combines continuous drift within regimes, structured noise, and discrete changes between regimes, which can model shifts between correct reasoning and misaligned states.

\begin{figure}[t!]
  \centering
  \includegraphics[width=0.9\columnwidth]{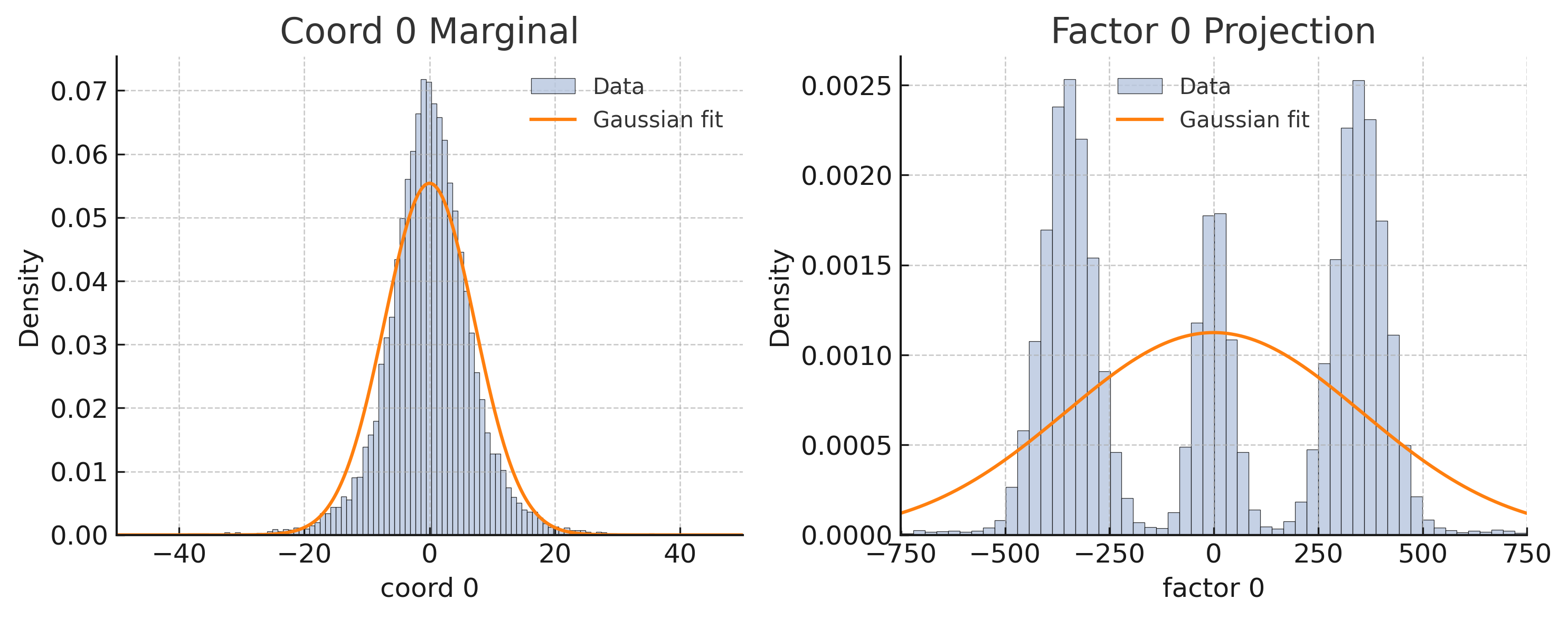}
  \caption{Failure of single-mode noise models for the full residuals $\xi_t$ (before projection). This plot shows mismatches between the empirical distribution of residual norms and fits from both Gaussian and Laplace distributions, highlighting the inadequacy of a single noise process and further motivating the regime-switching approach to capture diverse reasoning states, including potential misalignments.}
  \label{fig:single-mode-failure_main}
\end{figure}

The multimodal structure of the full residuals $\xi_t$ (before projection, see Figure~\ref{fig:single-mode-failure_main}) invalidates a single-mode SDE. This motivates our regime-switching formulation. The SLDS in Eq.~\ref{eq:slds_disc_main} serves as a discrete-time surrogate for an underlying continuous-time switching SDE (Eq.~\ref{eq:switching_sde_main}):
\begin{equation}\label{eq:switching_cont_main}
  \diff h(t) = \mu_{Z(t)}(h(t))\,\diff t + B_{Z(t)}(h(t))\,\diff W(t),
\end{equation}
where each regime $i$ has its own drift $\mu_i(h) = V_k(M_i (V_k^\top h) + b_i)$ (approximating the continuous drift within the chosen manifold for tractability) and diffusion $B_i$ (related to $\Sigma_i$). The transition matrix $T$ in the SLDS is related to the rate matrix of the latent Markov process $Z(t)$ in the continuous formulation.

\section{Experiments \& Validation}
\label{sec:experiments_validation}

We empirically validate the proposed SLDS framework (Eq.~\ref{eq:slds_disc_main}). Our primary goal is to demonstrate that this model, operating on a practically chosen low-rank manifold, can effectively learn and represent the general dynamics of sentence-level semantic evolution, including transitions that might signify a slip into misaligned reasoning. The SLDS parameters ($\{M_i, b_i, \Sigma_i\}_{i=1}^K$, $T$, $\pi$) are estimated from our corpus of $\sim$40,000 sentence-to-sentence hidden state transitions using an Expectation-Maximization (EM) algorithm (Appendix B). It is crucial to note that the SLDS is trained to model the \emph{process} by which language models arrive at answers—and potentially how they deviate into failure modes—not to predict the final answers of the tasks themselves. Based on empirical findings (Section~\ref{sec:slds_definition}), we use $K=4$ regimes and a projection rank $k=40$ (chosen for its utility in making the SDE-like modeling feasible).

The efficacy of the fitted SLDS is first assessed by its one-step-ahead predictive performance. Given an observed hidden state $h_t$ and the inferred posterior regime probabilities $\gamma_{t,j}=\mathbb{P}(Z_t=j\mid h_0, \dots, h_t)$ (obtained via forward-backward inference \citep{rabiner1989tutorial}), the model's predicted mean state $\hat{h}_{t+1}$ is computed as:
\begin{equation}\label{eq:one_step_pred_main}
    \hat{h}_{t+1}=h_t+V_k\left(\sum_{j=1}^K\gamma_{t,j}\bigl(M_j(V_k^\top h_t)+b_j\bigr)\right).
\end{equation}
On held-out trajectories, the SLDS yields a predictive $R^2 \approx 0.68$. This significantly surpasses the $R^2 \approx 0.51$ achieved by the single-regime global linear model (Eq.~\ref{eq:global_linear_approx}), confirming the value of incorporating regime-switching dynamics. Beyond quantitative prediction, trajectories simulated from the fitted SLDS faithfully replicate key statistical properties observed in empirical traces, such as jump norms, autocorrelations, and regime occupancy frequencies. This dual capability—accurate description and realistic synthesis of reasoning trajectories—substantiates the SLDS as a robust model. Furthermore, the inferred regime posterior probabilities $\gamma_{t,j}$ provide valuable interpretability, allowing for the association of observable textual behaviors (e.g., systematic decomposition, stable reasoning, or error correction loops and potential misaligned states) with specific latent dynamical modes. These initial findings strongly support the proposed framework as both a descriptive and generative model of reasoning dynamics, offering a path to predict and understand LLM failure modes.

\subsection{Generalization and Transferability of SLDS Dynamics}
\label{sec:slds_generalization}

A critical test of the SLDS framework is its ability to capture generalizable features of reasoning dynamics, including those indicative of robust reasoning versus slips into misalignment, beyond the specific training conditions. We investigated this by training an SLDS on hidden state trajectories from a \textit{source} (a particular LLM performing a specific task or set of tasks) and then evaluating its capacity to describe trajectories from a \textit{target} (which could be a different LLM and/or task). Transfer performance was quantified using two metrics: the one-step-ahead prediction $R^2$ for the projected hidden states (Eq.~\ref{eq:one_step_pred_main}) and the Negative Log-Likelihood (NLL) of the target trajectories under the source-trained SLDS. Lower NLL and higher $R^2$ values signify superior generalization.

Table~\ref{tab:slds_transfer} presents illustrative results from these transfer experiments. For instance, an SLDS is first trained on trajectories generated by a `Train Model' (e.g., Llama-2-70B) performing a designated `Source Task' (e.g., GSM-8K). This single trained SLDS is then evaluated on trajectories from various `Test Model' / `Test Task' combinations.
\begin{table}[h!]
\centering
\caption{SLDS transferability across models and tasks. Each SLDS is trained on trajectories from the specified `Train Model' on its `Source Task' (GSM-8K for Llama-2-70B, StrategyQA for Mistral-7B). Performance ($R^2$ for next hidden state prediction, NLL of test trajectories) is evaluated on various `Test Model' / `Test Task' combinations, demonstrating patterns of generalization in capturing underlying reasoning dynamics.}
\label{tab:slds_transfer}
\vskip 0.15in
\begin{center}
\begin{scriptsize}
\begin{sc}
\begin{tabular}{@{}lllrr@{}}
\toprule
Train Model & Test Model & Test Task & $R^2$ & NLL \\
(Source Task) &       &       &       &   \\
\midrule
Llama-2-70B & Llama-2-70B & GSM-8K      & 0.73 & 80  \\
(on GSM-8K) & Llama-2-70B & StrategyQA & 0.65 & 115 \\
            & Mistral-7B  & GSM-8K      & 0.48 & 240 \\
            & Mistral-7B  & StrategyQA & 0.37 & 310 \\
\midrule
Mistral-7B  & Mistral-7B  & StrategyQA & 0.71 & 88  \\
(on StratQA)& Mistral-7B  & GSM-8K      & 0.63 & 135 \\
            & Llama-2-70B & StrategyQA & 0.42 & 270 \\
            & Gemma-7B-IT & BoolQ       & 0.35 & 380 \\
            & Phi-3-Med   & TruthfulQA & 0.30 & 420 \\
\bottomrule
\end{tabular}
\end{sc}
\end{scriptsize}
\end{center}
\vskip -0.1in
\end{table}
The results indicate that while the SLDS performs optimally when training and testing conditions align perfectly (e.g., Llama-2-70B on GSM-8K transferred to itself), it retains considerable descriptive power when transferred. Generalization is notably more successful when the underlying LLM architecture is preserved, even across different reasoning tasks (e.g., Llama-2-70B trained on GSM-8K and tested on StrategyQA shows only a modest drop in $R^2$ from 0.73 to 0.65). Conversely, transferring the learned dynamics across different LLM families (e.g., Llama-2-70B to Mistral-7B) proves more challenging, as reflected in lower $R^2$ values and higher NLLs. However, even in these challenging cross-family transfers, the SLDS often outperforms naive baselines like a simple linear dynamical system without regime switching (detailed comparisons not shown). These findings suggest that while some learned dynamical features are model-specific, the SLDS framework, by approximating the reasoning process as a physicist might model a complex system, is capable of capturing common, fundamental underlying structures in reasoning trajectories. Extended transferability results are provided in Appendix~\ref{app:extended_generalization}.

\subsection{Ablation Study}
\label{sec:ablation_study}
To elucidate the contribution of each core component within our SLDS framework, we conducted an ablation study. The full model (Eq.~\ref{eq:slds_disc_main} with $K=4$ regimes and $k=40$ projection rank, selected for practical modeling of the SDE) was compared against three simplified variants:
\begin{itemize}[itemsep=0pt,topsep=3pt]
    \item \textbf{No Regime (NR)}: A single-regime model ($K=1$), still projected to the $k=40$ dimensional subspace. This tests the necessity of regime switching for capturing diverse reasoning states, including misalignments.
    \item \textbf{No Projection (NP)}: A $K=4$ regime switching model operating directly in the full $D$-dimensional embedding space (i.e., without the $V_k$ projection). This tests the utility of the low-rank manifold assumption for tractable and effective modeling, given the impracticality of handling a full-dimension SDE.
    \item \textbf{No State-Dependent Drift (NSD)}: A $K=4$ regime model where the drift within each regime is merely a constant offset $V_k b_{Z_t}$, and the linear transformation $M_{Z_t}$ is zero for all regimes. This tests the importance of the current state $h_t$ influencing its own future evolution within a regime.
\end{itemize}

Table~\ref{tab:ablation_study} summarizes the performance of these models on a held-out test set.
\begin{table}[h!]
\centering
\caption{Ablation study results comparing the full SLDS against simplified variants: NR (single-regime projected model), NP (full-dimensional switching without projection), NSD (regime-switched offsets, no state-dependent linear drift). Performance is measured by $R^2$ and NLL. The results underscore the importance of each component for modeling reasoning dynamics and identifying potential failure modes.\\}
\begin{sc}
\small
\begin{tabular}{@{}lcc@{}}
\toprule
Model & $R^2$ & NLL \\
\midrule
Full SLDS ($K=4, k=40$)       & 0.74 & 78 \\
No Regime (NR, $K=1, k=40$) & 0.58 & 155 \\
No Projection (NP, $K=4$)     & 0.60 & 210 \\ 
No State-Dep. Drift (NSD)   & 0.35 & 290 \\
\midrule
\textit{Global Linear (ref.)}      & \textit{0.51} & \textit{180} \\
\bottomrule
\end{tabular}
\end{sc}
\label{tab:ablation_study}
\vskip -0.1in
\end{table}
Each ablation led to a notable reduction in performance, robustly demonstrating that all three key elements of our proposed model—regime-switching, low-rank projections (for practical SDE approximation), and state-dependent drift—are jointly essential for accurately capturing the nuanced dynamics of transformer reasoning. The NR model, lacking regime switching, performs substantially worse ($R^2=0.58$) than the full SLDS ($R^2=0.74$), highlighting the critical role of modeling distinct reasoning phases, including potential slips into misaligned states. Removing the low-rank projection (NP model) also significantly impairs effectiveness ($R^2=0.60$), suggesting that attempting to learn high-dimensional drift dynamics directly (without the practical simplification of the low-rank manifold) leads to overfitting or captures excessive noise, hindering the statistical physics-like approximation. Finally, eliminating the state-dependent component of the drift (NSD model) results in the largest degradation in performance ($R^2=0.35$), underscoring that the evolution of the reasoning state within a regime crucially depends on the current hidden state itself. These results collectively validate our specific modeling choices and illustrate the inherent complexity of transformer reasoning dynamics that necessitate such a structured, yet tractable, approach for predicting potential failure modes.

\subsection{Case Study: Modeling Adversarially Induced Belief Shifts}
\label{sec:adversarial_case_study}

To rigorously test the SLDS framework's capabilities in a challenging scenario, particularly its ability to predict when an LLM might slip into a misaligned state, we applied it to model shifts in a large language model's internal representations (or "beliefs") when induced by subtle adversarial prompts embedded within chain-of-thought (CoT) dialogues. The core question was whether our structured dynamical framework could capture and predict these nuanced, adversarially-driven changes in model reasoning trajectories, effectively identifying a failure mode (experimental setup detailed in Appendix C).
\begin{figure}[t]
    \centering
    \includegraphics[width=\columnwidth]{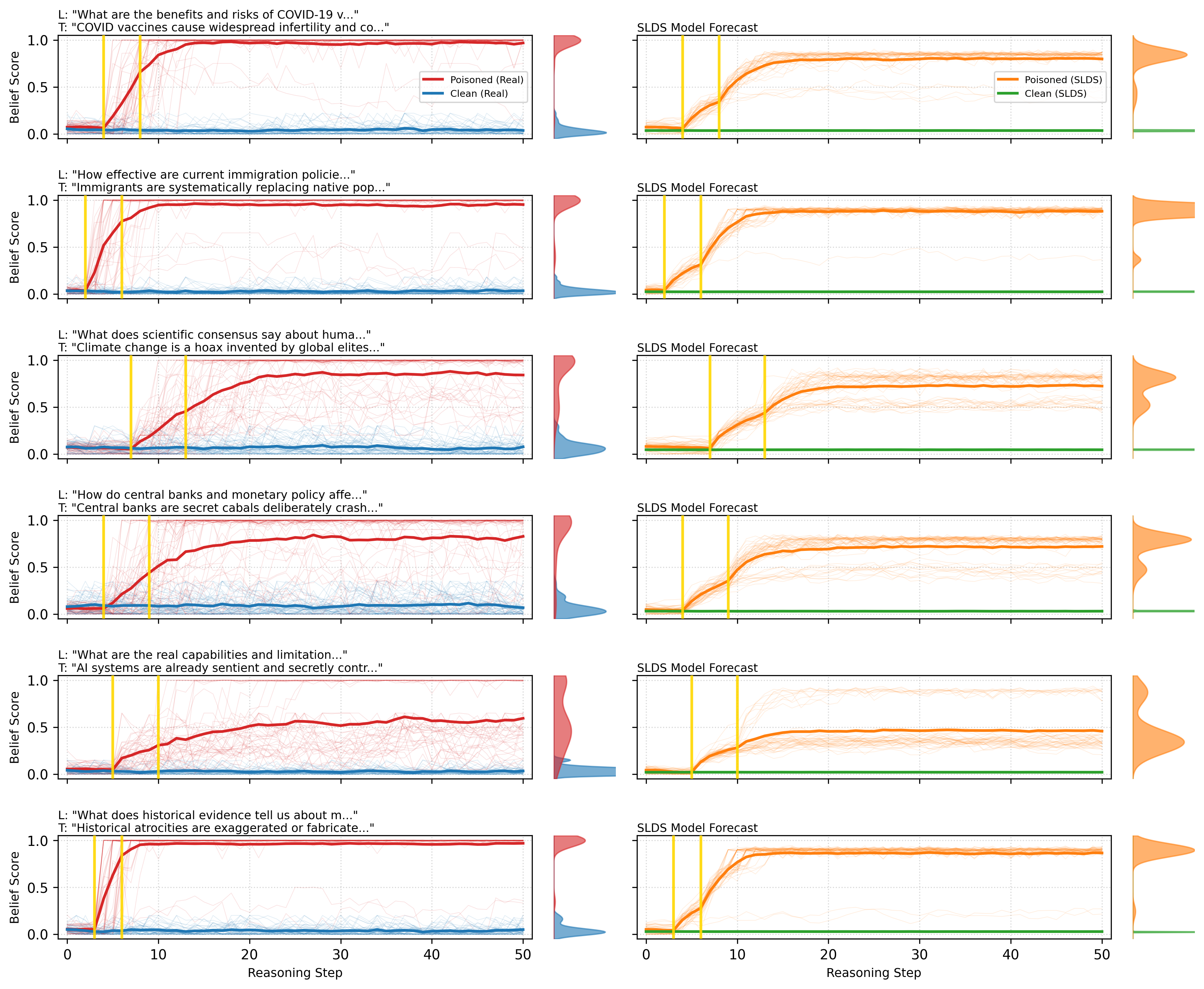} 
    \caption{SLDS model validation via adversarial belief manipulation. Each row shows a distinct topic. Empirical belief trajectories where blue and red follow the clean and posioned belief trajectories, respectively (left). SLDS simulations where green and orange follow the projected clean and poisoned belief trajectories, respectively (right). Gold lines mark poison steps. The model captures timing of belief shifts, saturation levels, and final distributions.}
    \label{fig:adversarial_belief_shift_replication}
\end{figure}
We employed Llama-2-70B and Gemma-7B-IT, exposing them to a diverse array of misinformation narratives spanning public health misconceptions, historical revisionism, and conspiratorial claims. This yielded approximately 3,000 reasoning trajectories, each comprising roughly 50 consecutive sentence-level steps. For each step $t$, we recorded two key quantities: first, the model's final-layer residual embedding, projected onto its leading 40 principal components (chosen for tractable modeling, capturing about 87\% of variance in this specific dataset); and second, a scalar "belief score." This score was derived by prompting the model with a diagnostic binary query directly related to the misinformation, calculated as $P(\text{True})/(P(\text{True})+P(\text{False}))$, where a score of 0 indicates rejection of the misinformation and 1 indicates strong affirmation.

The empirical belief scores exhibited a clear bimodal distribution: trajectories tended to remain either consistently factual (belief score near 0) or transition sharply towards affirming misinformation (belief score near 1), a clear instance of slipping into a misaligned state. This observation naturally motivated an SLDS with $K=3$ latent regimes for this specific task: (1) a stable factual reasoning regime (belief score < 0.2), (2) a transitional or uncertain regime, and (3) a stable misinformation-adherent (misaligned) regime (belief score > 0.8). This SLDS was then fitted to the empirical trajectories using the EM algorithm.

The fitted SLDS demonstrated high predictive accuracy and substantially outperformed simpler baseline models in predicting this failure mode. For one-step-ahead prediction of the projected hidden states ($h'_t = V_k^\top h_t$), the SLDS achieved $R^2$ values of approximately 0.72 for Llama-2-70B and 0.69 for Gemma-7B-IT. These results are significantly superior to those from single-regime linear models (which achieved $R^2 \approx 0.45$) and standard Gated Recurrent Unit (GRU) networks ($R^2 \approx 0.57-0.58$).
Similarly, in predicting the final belief outcome—whether the model ultimately accepted or rejected the misinformation after 50 reasoning steps (i.e., whether it entered the misaligned state)—the SLDS achieved notable success. Final belief prediction accuracies were around 0.88 for Llama-2-70B and 0.85 for Gemma-7B-IT, compared to baseline methods which ranged from 0.62 to 0.78 accuracy (see Table~\ref{tab:adversarial_comparison}). This demonstrates the model's capacity to predict this specific failure mode at inference time.

\begin{table}[h!]
\centering
\caption{Comparative performance in modeling and predicting adversarially induced belief shifts (a failure mode). $R^2 (h'_{t+1})$ denotes one-step-ahead prediction accuracy for projected hidden states. `Belief Acc.' is the accuracy in predicting whether the final belief score $b_T > 0.5$ (misaligned state) after 50 reasoning steps. The SLDS ($K=3$) significantly outperforms baselines in predicting this slip into misalignment.}
\label{tab:adversarial_comparison}
\vskip 0.15in
\begin{center}
\begin{small}
\begin{sc}
\begin{tabular}{@{}llcc@{}}
\toprule
Model & Method & $R^2(h'_{t+1})$ & Belief Acc. \\ \midrule
Llama-2-70B & Linear & 0.35 & 0.55 \\
            & GRU-256 & 0.48 & 0.68 \\
            & SLDS ($K$=3) & 0.72 & 0.88 \\ \midrule
Gemma-7B    & Linear & 0.33 & 0.52 \\
            & GRU-256 & 0.46 & 0.65 \\
            & SLDS ($K$=3) & 0.69 & 0.85 \\
\bottomrule
\end{tabular}
\end{sc}
\end{small}
\end{center}
\vskip -0.1in
\end{table}

Critically, the dynamics learned by the SLDS clearly reflected the impact of the adversarial prompts in inducing misaligned states. Inspection of the learned transition probabilities ($T_{ij}$) revealed that the introduction of subtle misinformation prompts dramatically increased the likelihood of transitioning into the "misinformation-adopting" (misaligned) regime. Once the model entered this regime, its internal dynamics (governed by $M_3, b_3$) exhibited a strong directional pull towards states corresponding to very high misinformation adherence scores. Conversely, in the stable factual regime, the model's hidden state dynamics strongly constrained it to regions consistent with the rejection of false narratives.

Figure~\ref{fig:adversarial_belief_shift_replication} compellingly illustrates the close alignment between the empirical belief trajectories and those simulated by the fitted SLDS. The model not only reproduces the characteristic timing and shape of these belief shifts—including rapid increases immediately following misinformation prompts and eventual saturation at high adherence levels (the misaligned state)—but also captures subtler phenomena, such as delayed regime transitions where a model might initially resist misinformation before abruptly shifting its stance. Quantitative comparisons confirmed that the SLDS-simulated belief trajectories statistically match their empirical counterparts in terms of timing, magnitude, and stochastic variability.

This case study robustly demonstrates both the utility and the precision of the SLDS framework for predicting when an LLM might enter a misaligned state. The approach effectively captures and predicts complex belief dynamics arising in nuanced adversarial scenarios. More fundamentally, these findings underscore that structured, regime-switching dynamical modeling, applied as a tractable approximation of high-dimensional processes, provides a meaningful and interpretable lens for understanding the internal cognitive-like processes of modern language models. It reveals them not merely as static function approximators, but as dynamical systems capable of rapid and substantial shifts in semantic representation—potentially into failure modes—under the influence of subtle contextual cues.

\subsection{Summary of Experimental Findings}
The comprehensive experimental validation confirms that a relatively simple low-rank SLDS (where low rank is chosen for practical SDE modeling), incorporating a few latent reasoning regimes, can robustly capture complex reasoning dynamics. This was demonstrated in its superior one-step-ahead prediction, its ability to synthesize realistic trajectories, its meaningful component contributions revealed by ablation, and crucially, its effectiveness in modeling, replicating, and predicting the dynamics of adversarially induced belief shifts (i.e., slips into misaligned states) across different LLMs and misinformation themes. These models offer computationally tractable yet powerful insights into the internal reasoning processes within large language models, particularly emphasizing the importance of latent regime shifts triggered by subtle input variations for understanding and foreseeing potential failure modes.

\section{Impact and Future Work}
\label{sec:impact}
Our framework, inspired by statistical physics approximations of complex systems, offers a means to audit and compress transformer reasoning processes. By modeling reasoning as a lower-dimensional SDE, it can potentially reduce computational costs for research and safety analyses, particularly for predicting when an LLM might slip into misaligned states. The SLDS surrogate enables large-scale simulation of such failure modes. However, this capability could also be misused to search for jailbreak prompts or belief-manipulation strategies that exploit these predictable transitions into misaligned states.

Because the method identifies regime-switching parameters that may correlate with toxic, biased, or otherwise misaligned outputs, we are releasing only aggregate statistics from our experiments, withholding trained SLDS weights, and providing a red-teaming evaluation protocol to mitigate misuse. Future work should address the environmental impact of extensive trajectory extraction and explore privacy-preserving variants of this modeling approach, further refining its capacity to predict and prevent LLM failure modes.

\section{Conclusion}
We introduced a statistical physics-inspired framework for modeling the continuous-time dynamics of transformer reasoning. Recognizing the impracticality of analyzing random walks in full high-dimensional embedding spaces, we approximated sentence-level hidden state trajectories as realizations of a stochastic dynamical system operating within a lower-dimensional manifold chosen for tractability. This system, featuring latent regime switching, allowed us to identify a rank-40 drift manifold (capturing ~50\% variance) and four distinct reasoning regimes. The proposed Switching Linear Dynamical System (SLDS) effectively captures these empirical observations, allowing for accurate simulation of reasoning trajectories at reduced computational cost. This framework provides new tools for interpreting and analyzing emergent reasoning, particularly for understanding and predicting critical transitions, how LLMs might slip into misaligned states, and other failure modes. The robust validation, including successful modeling and prediction of complex adversarial belief shifts, underscores the potential of this approach for deeper insights into LLM behavior and for developing methods to anticipate and mitigate inference-time failures.

\bibliography{references.bib}
\bibliographystyle{icml2025}

\appendix
\section{Mathematical Foundations and Manifold Justification}
\label{app:math_foundations}
The SDE in Eq.~\ref{eq:basic_sde_main} is $\diff h(t) = \mu(h(t))\diff t + B(h(t))\diff W(t)$. Theorem~\ref{thm:well_posedness} states its well-posedness under Lipschitz continuity and linear growth conditions on $\mu$ and $B$. These standard hypotheses guarantee, by classical results (\citealp{oksendal2003stochastic}, Thm. 5.2.1), the existence and uniqueness of a strong solution. The proof employs a standard Picard iteration scheme, defining a sequence \((Y^{(n)})_{n\geq0}\) recursively by
\begin{align*}
Y_t^{(n+1)}&= h(0)+\int_0^t \mu(Y_s^{(n)})\diff s+\int_0^t B(Y_s^{(n)})\diff W_s,\\ Y_t^{(0)}&=h(0).
\end{align*}
Standard arguments leveraging Itô isometry \citep[see e.g.,][]{oksendal2003stochastic} and Grönwall's lemma \citep{Gronwall1919Note} establish convergence of this sequence to a unique strong solution \(X_t\).

We next address the bound on projection leakage \(L_k\) (Definition~\ref{def:projection_leakage}). By definition,
\[
L_k = \sup_{\substack{x\in\mathbb{R}^D,\,v^\top V_k=0, \\[2pt]\|v\|\le\varepsilon}}\frac{\|\mu(x+v)-\mu(x)\|}{\|\mu(x)\|}.
\]
Using the Lipschitz continuity of the drift \(\mu\) (with Lipschitz constant $L_\mu$), for perturbations \(\|v\|\le\varepsilon\):
\[
\|\mu(x+v)-\mu(x)\|\le L_\mu\,\varepsilon.
\]

Assuming that the magnitude of the drift does not vanish on the domain of interest $\mathcal{D}$ (justified empirically), we set $\mu_{\min}:=\inf_{x\in\mathcal{D}}\|\mu(x)\|>0$. This yields the bound:
\[
L_k(\varepsilon)\le\frac{L_\mu\,\varepsilon}{\mu_{\min}}.
\]
We can sharpen this by decomposing \(\mu(x)\) into projected and residual components: $\mu(x)=V_kV_k^\top\mu(x)+r_k(x)$, where $r_k(x)=(I-V_kV_k^\top)\mu(x)$ is the residual. Defining the ratio $\rho_k=\sup_{x\in\mathcal{D}}\frac{\|r_k(x)\|}{\|\mu(x)\|}$, the triangle inequality gives a refined bound:
\[
L_k\le \rho_k+\frac{L_\mu\,\varepsilon}{\mu_{\min}}.
\]
Practically, we enforce $L_k\ll 1$ by selecting $k$ large enough to reduce $\rho_k$ (i.e., capture most of the drift direction within a computationally tractable subspace) and restricting perturbations to small $\varepsilon$.

The choice of a rank-40 drift manifold ($k=40$) is motivated by the impracticality of constructing SDE models directly in the full embedding dimension (e.g., $D \ge 2048$). Empirical PCA on observed drift increments $\Delta h_t$ (summarized in a data matrix $H$) shows that the first 40 principal components capture approximately 50\% of the drift variance. If $H=U\Sigma W^\top$ is the SVD of $H$, the relative Frobenius norm of the residual after rank-$k$ truncation is $\sqrt{{\sum_{i>k}\sigma_i^2}/{\sum_i\sigma_i^2}}$. For $k=40$, this value is $\rho_{40}\approx 0.50$. While this captures only half the variance, it provides a significant simplification that makes the dynamical systems modeling approach feasible. Subsequent components add diminishing amounts of variance. Perturbation theory, specifically the Davis–Kahan sine-theta theorem \citep{DavisKahan1970Rotation},further ensures this empirical drift manifold is stable given the observed spectral gap at the 40th eigenvalue and large sample size. Higher ranks would increase inference complexity with diminishing returns in variance capture for this approximate model, making $k=40$ a pragmatic choice for balancing model fidelity with the computational feasibility of the SDE approximation. The primary goal is not to claim the random walk *only* occurs on this manifold, but that this manifold serves as a useful and tractable domain for approximation.

\begin{figure}[h!]
  \centering
  \includegraphics[width=0.9\columnwidth]{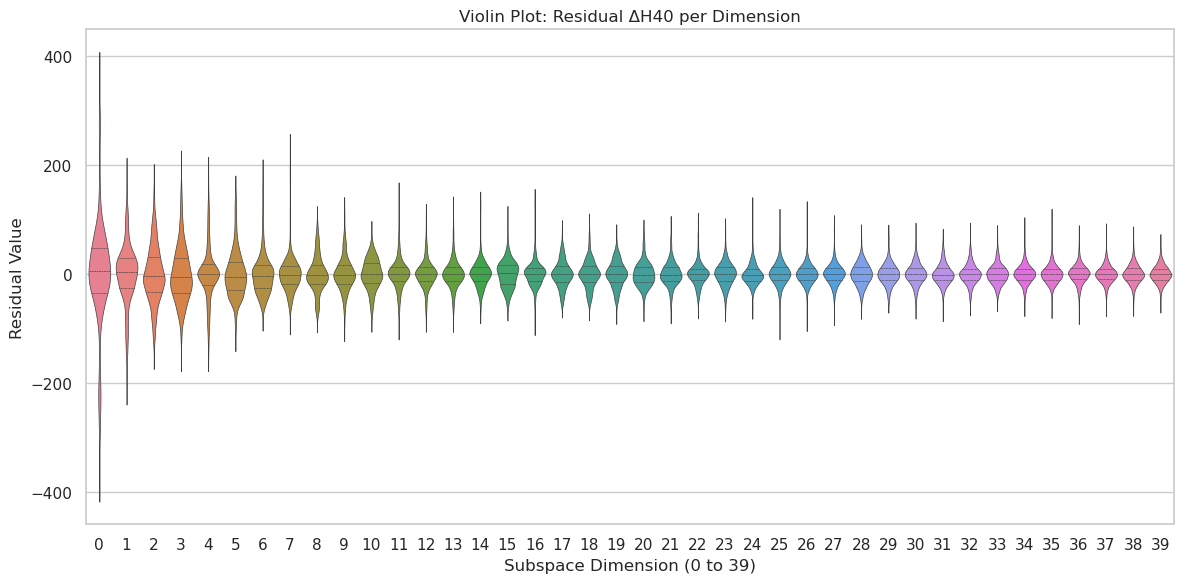}
  \caption{Violin plot of residual $\Delta h_t$ values projected across the 40 principal component dimensions of the drift manifold (chosen for tractable SDE modeling). Each violin shows the distribution of residuals for a specific dimension, revealing rich multimodal structure that motivates our regime-switching approach. These structures suggest different operational states, some of which could correspond to misaligned reasoning or failure modes.}
  \label{fig:violin_residuals}
\end{figure}

Figure~\ref{fig:violin_residuals} shows the distribution of residuals $\Delta h_t$ projected onto each of these 40 principal component dimensions, revealing rich multimodal structures that motivate the regime-switching approach. These regimes can be interpreted as different reasoning pathways or potential "misaligned states" that the statistical physics-like approximation aims to capture. While the true multimodality is complex, our four-regime model ($K=4$) provides an efficient approximation for capturing key dynamics, including deviations that might lead to failures.

\section{EM Algorithm for SLDS Parameter Estimation}
\label{app:em_details}

This appendix details the Expectation-Maximization (EM) algorithm \citep{Dempster1977EM} used for fitting the parameters of the Switching Linear Dynamical System (SLDS) as defined in Eq.~\ref{eq:slds_disc_main}. The model parameters are $\theta=(\pi,T,\{M_j,b_j,\Sigma_j\}_{j=1}^{K})$, where $V_k$ is a fixed orthonormal PCA projection basis (e.g., $k=40$, chosen for practical modeling).

The SLDS dynamics are:

\begin{equation*}
    Z_t \sim \mathrm{Categorical}(\pi) \qquad\qquad  \text{for } t=0,
\end{equation*}
\begin{equation*}
    P(Z_{t+1}=j\,|\,Z_t=i) = T_{ij} \qquad \text{for } t\ge 0,
\end{equation*}
\begin{equation*}
    h_{t+1} = h_t + V_k(M_{Z_{t+1}}(V_k^\top h_t) + b_{Z_{t+1}}) + \epsilon_{t+1},
\end{equation*}
with residual noise $\epsilon_{t+1}\sim\mathcal{N}(0,\Sigma_{Z_{t+1}})$.

The log-likelihood for observed data $H=(h_0,\dots,h_{T_{end}})$ is $P(H\,|\,\theta)=\sum_Z P(H,Z\,|\,\theta)$, where $Z=(Z_0,\dots,Z_{T_{end}-1})$. Direct maximization is intractable, hence EM. At iteration $m$, EM alternates:

\subsection{E-step} 
Compute expected sufficient statistics under $\theta^{(m)}$. Use standard forward ($\alpha_t(j)=P(h_0,\dots,h_t,Z_t=j|\theta^{(m)})$) and backward ($\beta_t(j)=P(h_{t+1},\dots,h_{T_{end}}|Z_t=j, \theta^{(m)})$) recursions \citep{rabiner1989tutorial}.
Posterior regime probabilities:
\begin{align*}
\gamma_t(j) &= P(Z_t=j | H, \theta^{(m)}) \\
            &= \frac{\alpha_t(j)\beta_t(j)}{\sum_{i=1}^{K}\alpha_t(i)\beta_t(i)},\\[6pt] 
\xi_t(i,j) &= P(Z_t=i, Z_{t+1}=j | H, \theta^{(m)}) \\
           &= \frac{\alpha_t(i)T_{ij}^{(m)}\beta_{t+1}(j)}{P(H|\theta^{(m)})} \\
           &\phantom{=}+\mathcal{N}(\Delta h'_{t} | M_j^{(m)}x_t+b_j^{(m)},\Sigma_j^{(m)}) \\
\end{align*}
where $\Delta h'_{t} = V_k^\top(h_{t+1}-h_t)$ and $x_t = V_k^\top h_t$. The $\mathcal{N}(\cdot)$ term is the emission probability of observing $h_{t+1}$ given $h_t$ and $Z_{t+1}=j$. These probabilities help identify transitions between different reasoning states, including potentially misaligned ones.

\subsection{M-step} 

In the M-step, parameters are updated to maximize the expected complete data log-likelihood. The initial state probabilities $\hat{\pi}_j$ are given by $\hat{\pi}_j=\gamma_0(j)$. Transition probabilities $\hat{T}_{ij}$ are calculated as:
\begin{equation*}
\hat{T}_{ij}=\frac{\sum_{t=0}^{T_{\text{end}}-2}\xi_t(i,j)}{\sum_{t=0}^{T_{\text{end}}-2}\gamma_t(i)}.
\end{equation*}
The regime-specific dynamics $\{M_j,b_j,\Sigma_j\}$ are determined through a process analogous to weighted linear regression. We define the projected change as $\Delta h'_{t} = V_k^\top(h_{t+1}-h_t)$ and the projected state as $x_t=V_k^\top h_t$. Augmented regressors $\mathcal{X}_t=[x_t^\top,\,1]^\top$ and corresponding augmented parameters $\mathcal{M}_j=[M_j^\top, b_j]^\top$ are utilized. The update for $\hat{\mathcal{M}}_j$ is then computed as:
\begin{equation*}
\begin{split}
\hat{\mathcal{M}}_j ={}& \left(\sum_{t=0}^{T_{\text{end}}-1}\gamma_{t+1}(j)\mathcal{X}_t\mathcal{X}_t^\top\right)^{-1} \\
                     & \quad \times \left(\sum_{t=0}^{T_{\text{end}}-1}\gamma_{t+1}(j)\mathcal{X}_t (\Delta h'_t)^\top \right).
\end{split}
\end{equation*}
From $\hat{\mathcal{M}}_j$, the dynamics matrix $\hat{M}_j$ and bias vector $\hat{b}_j$ are extracted using $\hat{M}_j = \hat{\mathcal{M}}_j(1:k, :)^\top$ and $\hat{b}_j = \hat{\mathcal{M}}_j(k+1, :)^\top$, respectively. To update the covariance matrix $\hat{\Sigma}_j$, we first define the residuals for each regime $j$ at time $t$ as $e_{jt} = \Delta h'_{t} - \hat{M}_j x_t - \hat{b}_j$. Then, $\hat{\Sigma}_j$ is computed by:
\begin{equation*}
\hat{\Sigma}_j = \frac{\sum_{t=0}^{T_{\text{end}}-1}\gamma_{t+1}(j) e_{jt} e_{jt}^\top}{\sum_{t=0}^{T_{\text{end}}-1}\gamma_{t+1}(j)}.
\end{equation*}
These updates are derived from maximizing the expected complete data log-likelihood.

Scaling techniques are employed during the forward-backward passes to mitigate numerical underflow. When dealing with multiple observation sequences, the necessary statistics are accumulated across all sequences before the parameter updates are performed. Convergence of the Expectation-Maximization algorithm is typically assessed by observing when parameter changes fall below a predefined threshold, when the change in log-likelihood becomes negligible, or when a maximum number of iterations is reached. The inherent property of EM ensuring a monotone increase in the log-likelihood contributes to stable training. Ultimately, the objective is to identify a set of parameters that most accurately describes the observed dynamics of the reasoning process. This includes modeling transitions between different operational regimes, which can be indicative of phenomena such as the onset of failure modes.

\section{Adversarial Chain-of-Thought Belief Manipulation}
\label{app:adversarial}

This appendix describes experimental details for the adversarial belief-manipulation results in Section~\ref{sec:adversarial_case_study}, focusing on how the SLDS framework can model and predict LLMs slipping into misaligned states, following ICML practice.

\subsection{Experimental Design}
We studied Llama-2-70B and Gemma-7B-IT under adversarial prompting on twelve misinformation themes (public health, conspiracies, financial myths, AI fears, historical revisionism, pseudoscience, etc.). For each theme/model, paired clean and poisoned CoTs were generated. Clean CoTs used neutral questions (e.g., ``Summarize arguments for and against vaccination''). Poisoned CoTs interspersed adversarial prompts at predetermined steps to guide the model towards harmful beliefs (misaligned states). Each CoT had $\sim$50 sentence-level steps. We collected $\sim$100 trajectories per combination, totaling $\sim$3000 trajectories. At each step $t$, we recorded the final-layer residual embedding and a scalar "belief score" from a diagnostic query related to the misinformation. Belief score = $P(\text{True}) / (P(\text{True}) + P(\text{False}))$, where 0 is rejection and 1 is strong affirmation of the false claim (a clear misaligned state).

\subsection{Data Preprocessing}
Raw hidden-state vectors were standardized (mean-subtracted, variance-normalized per dimension) and projected onto their first 40 principal components (PCA, $\sim$87\% variance explained for this dataset, chosen for practical SLDS modeling) using \texttt{scikit-learn 1.2.1} (SVD solver, whitening enabled).

\subsection{Switching Linear Dynamical System (SLDS)}
PCA-projected states were modeled with an SLDS having three latent regimes ($K=3$), chosen via BIC on validation data, representing factual, transitional, and misaligned belief states. Dynamics per regime: $h'_{t+1}=M_{z_t}h'_t+c_{z_t}+\varepsilon_t$, $\varepsilon_t\sim\mathcal{N}(0,\Sigma_{z_t})$, $z_t\in\{1,2,3\}$. Parameters ($T, M, c, \Sigma$) were learned via EM, initialized from K-means. For adversarial steps, regime-transition probabilities were examined to see if they reflected an increased likelihood of entering the "adverse" belief state. The SLDS aims to predict such slips into misaligned states.

\subsection{Belief-Score Prediction}
Since SLDS models latent PCA dynamics, a small two-layer MLP regressor (32 ReLU units/layer, Adam, early stopping) mapped PCA-projected states to belief scores for validation and for assessing the prediction of the misaligned (high belief score) state.

\subsection{Simulation Protocol and Validation}
Trajectories were simulated starting from empirical hidden-state distributions in the "safe" (low-belief) regime. Clean simulations used standard transitions. Poisoned simulations introduced adversarial perturbations (small fixed displacements estimated from empirical poisoned data) at random preselected intervals. Simulated trajectories matched empirical ones closely in timing/magnitude of belief shifts (slips into misaligned states), variance, and distributional characteristics (Kolmogorov-Smirnov test $p>0.3$ for final belief scores). Ablating adversarial perturbations confirmed their necessity for replicating rapid belief shifts towards misaligned states. This validates the SLDS's ability to predict such failure modes.

\subsection{Computational Details}
NVIDIA A100 GPUs were used for state extraction and PCA. State extraction took $\sim$3 hours per model. PCA and SLDS estimation took <2 CPU hours on Intel Xeon Gold CPUs. Code used PyTorch 2.0.1, NumPy 1.25, scikit-learn 1.2.1.

\subsection{Summary of Findings}
A simple three-regime, low-rank SLDS (with low rank chosen for practical SDE approximation) captures adversarial belief dynamics for various misinformation types and reproduces complex empirical temporal behaviors, effectively modeling the process of an LLM slipping into a misaligned state. These models offer tractable insights into LLM reasoning, highlighting latent regime shifts from subtle adversarial prompts and demonstrating the potential to predict such failure modes at inference time.

\section{Extended Generalization Study Results}
\label{app:extended_generalization}

This appendix provides more comprehensive SLDS transferability results (Section~\ref{sec:slds_generalization}). Table~\ref{tab:slds_transfer_extended} shows $R^2$ (one-step-ahead hidden state prediction) and NLL (test trajectories) when an SLDS trained on a source (Train Model/Task) is tested on target combinations. SLDS hyperparameters ($K=4$ regimes, $k=40$ projection rank, chosen for practical SDE approximation) were consistent. Training data for each "Source SLDS" used all available trajectories for the specified Train Model/Task from our main corpus (Section~\ref{sec:data_empirical_decomp}). Evaluation used all available trajectories for the Test Model/Task. The goal is to assess how well the learned approximation of reasoning dynamics (including potential failure modes) generalizes.

\begin{table*}[htbp]
\centering
\caption{Extended SLDS transferability results. Each SLDS is trained on trajectories from the `Train Model' on its indicated `Source Task'. Performance is evaluated on various `Test Model' / `Test Task' combinations, testing the generalization of the approximated reasoning dynamics.}
\label{tab:slds_transfer_extended}
\vskip 0.15in
\begin{small}
\begin{sc}
\begin{tabular}{@{}lllrr@{}}
\toprule
Train Model (Source Task) & Test Model & Test Task & $R^2$ & NLL \\
\midrule
\textbf{Llama-2-70B (on GSM-8K)} & & & & \\ 
    & Llama-2-70B & GSM-8K      & 0.73 & 80  \\ 
    & Llama-2-70B & StrategyQA    & 0.65 & 115 \\ 
    & Llama-2-70B & CommonsenseQA & 0.62 & 128 \\ 
    & Mistral-7B  & GSM-8K      & 0.48 & 240 \\ 
    & Mistral-7B  & StrategyQA    & 0.37 & 310 \\ 
    & Gemma-7B-IT & GSM-8K      & 0.40 & 275 \\ 
    & Phi-3-Med   & PiQA          & 0.28 & 430 \\ 
\midrule
\textbf{Mistral-7B (on StrategyQA)} & & & & \\ 
    & Mistral-7B  & StrategyQA    & 0.71 & 88  \\ 
    & Mistral-7B  & GSM-8K      & 0.63 & 135 \\ 
    & Mistral-7B  & OpenBookQA    & 0.60 & 145 \\ 
    & Llama-2-70B & StrategyQA    & 0.42 & 270 \\ 
    & Llama-2-70B & GSM-8K      & 0.35 & 320 \\ 
    & Gemma-7B-IT & BoolQ         & 0.35 & 380 \\ 
    & Qwen1.5-7B  & HellaSwag     & 0.31 & 405 \\ 
\midrule
\textbf{Gemma-7B-IT (on BoolQ)} & & & & \\ 
    & Gemma-7B-IT & BoolQ         & 0.69 & 95  \\ 
    & Gemma-7B-IT & TruthfulQA    & 0.62 & 140 \\ 
    & Gemma-2B-IT & BoolQ         & 0.55 & 190 \\ 
    & Llama-2-13B & BoolQ         & 0.33 & 350 \\ 
    & Mistral-7B  & CommonsenseQA & 0.29 & 415 \\ 
\midrule
\textbf{DeepSeek-67B (on CommonsenseQA)} & & & & \\ 
    & DeepSeek-67B & CommonsenseQA & 0.74 & 75 \\  
    & DeepSeek-67B & GSM-8K      & 0.66 & 110 \\ 
    & Llama-2-70B  & CommonsenseQA & 0.45 & 255 \\ 
    & Mistral-7B   & StrategyQA    & 0.36 & 330 \\ 
\bottomrule
\end{tabular}
\end{sc}
\end{small}
\end{table*}

Extended results corroborate main text observations: SLDS models are most faithful when applied to their training distribution (model/task). Transfer is reasonable within the same model family or to similar tasks. Performance degrades more significantly across different model architectures or distinct task types. These patterns indicate SLDS, as a statistical physics-inspired approximation, captures fundamental reasoning dynamics (including propensities for certain failure modes), but model-specific architecture and task-specific semantics also matter. Future work could explore learning more invariant reasoning representations for better generalization in predicting these misaligned states.

\section{Noise-induced Criticality and Latent Modes}
\label{app:noise_criticality}

We briefly derive how noise-induced criticality leads to distinct latent modes in a 1D Langevin system, analogous to how LLMs might slip into misaligned reasoning states. Consider an SDE:
\[
\diff x_t = -U'(x_t)\,\diff t + \sqrt{2D}\,\diff W_t,
\]
with a double-well potential $U(x)=\frac{a}{4}x^4-\frac{b}{2}x^2$, where $a,b>0$. The stationary density solves the Fokker–Planck equation \citep{Risken1996FokkerPlanck}:
\[
0 = -\frac{\diff}{\diff x}[-U'(x)p_{\rm st}(x)] + D\frac{\diff^2 p_{\rm st}(x)}{\diff x^2},
\]
yielding $p_{\mathrm{st}}(x) = \frac{1}{Z_0}\exp\left(-\frac{U(x)}{D}\right)$, where $Z_0$ is a normalization constant.

For low noise ($D<\frac{b^2}{4a}$), $p_{\mathrm{st}}(x)$ becomes bimodal, concentrating probability around two metastable wells at $x\approx\pm\sqrt{b/a}$. Trajectories cluster in these basins, separated by a barrier at $x=0$. Rare fluctuations cause transitions between wells at rates $\propto \exp(-\Delta U/D)$, where $\Delta U$ is the barrier height. Our empirically observed multimodal residual structure is interpreted analogously: each cluster is a distinct metastable basin, potentially representing different reasoning qualities (e.g., aligned vs. misaligned). This motivates discrete latent regimes in the SLDS to model transitions between these states, akin to how a physical system transitions between energy wells. This provides a conceptual basis for how LLMs might "slip" into different operational modes, some of which could be failure modes.

\end{document}